\documentclass[conference]{IEEEtran}
\IEEEoverridecommandlockouts
\usepackage{cite}
\usepackage{amsmath,amssymb,amsfonts}
\usepackage{algorithmic}
\usepackage{graphicx}
\usepackage{textcomp}
\usepackage[colorinlistoftodos]{todonotes}
\usepackage{xcolor}
\def\BibTeX{{\rm B\kern-.05em{\sc i\kern-.025em b}\kern-.08em
    T\kern-.1667em\lower.7ex\hbox{E}\kern-.125emX}}
\begin{document}

\title{Analyzing the Granularity and Cost of Annotation in Clinical Sequence Labeling}


\author{\IEEEauthorblockN{ Haozhan Sun}
\IEEEauthorblockA{
\textit{Duke University}\\
Durham, United States \\
haozhan.sun@duke.edu}
\and
\IEEEauthorblockN{Chenchen Xu}
\IEEEauthorblockA{
\textit{The Australian National University}\\
Canberra, Australia \\
chenchen.xu@anu.edu.au}
\and
\IEEEauthorblockN{Hanna Suominen}
\IEEEauthorblockA{
\textit{The Australian National University}\\
Canberra, Australia \\
hanna.suominen@anu.edu.au}}
\maketitle

\begin{abstract}
Well-annotated datasets, as shown in recent top studies, are becoming more important for researchers than ever before in supervised machine learning (ML). However, the dataset annotation process and its related human labor costs remain overlooked. In this work, we analyze the relationship between the annotation granularity and ML performance in sequence labeling, using clinical records from nursing shift-change handover. We first study a model derived from textual language features alone, without additional information based on nursing knowledge. We find that this sequence tagger performs well in most categories under this granularity. Then, we further include the additional manual annotations by a nurse, and find the sequence tagging performance remaining nearly the same. Finally, we give a guideline and reference to the community arguing it is not necessity and even not recommended to annotate in detailed granularity because of a low Return on Investment. Therefore we recommend emphasizing other features, like textual knowledge, for researchers and practitioners as a cost-effective source for increasing the sequence labeling performance.
\end{abstract}


\begin{IEEEkeywords}
Data processing; Electronic medical records; Health information management; Neural networks; Supervised learning; Text analysis
\end{IEEEkeywords}
\section{Introduction}

\noindent Natural Language Processing (NLP) techniques can derive a wide range of applications in medical and health practice. Specifically, in order to understand the clinical records well, we need deep understanding of the textual contents \cite{KREIMEYER201714,SpasicandGoran2020}. This yields the necessity for addressing downstream tasks under the concept of sequence labeling, which is our main focus on the clinical handover.

However, most commonly used and state-of-the-art NLP models are mostly supervised learning algorithms, which means the performance of the model is highly dependent on the quality and amount of labeled training data \cite{mitkov1999importance}. A well-annotated, high-quality dataset is regarded as the default prerequisite before researchers start to show their magic \cite{HovyandLavid2010,johnson-etal-2018-predicting}. Therefore, we argue that the analysis of the process of creating such annotated datasets is highly overlooked.

This work studies the process to automatically annotate clinical handover. In this process, we train a model using a labeled handover dataset such that the model can annotate future clinical records (i.e. identify which part is patient introduction and which part is the appointment information with a doctor).

In order to thoroughly study the data annotation process, the work mainly focuses on the relationship between the granularity of annotations and sequence labeling performance. Specifically, we try to answer \emph{``Is the cost of additional expert annotation worth it?"} Hence, we try different granularity levels when annotating and/or labeling the tags in a clinical handover dataset in order to analyze the difference in terms of the sequence labeling performance. We also study the ``return on investment" when detailing the granularity against the performance gain of the model. This is because of the nature of (clinical)text annotation --- the high cost of human labor resources.

In the end, we quantitatively formulate the human labor cost against the performance gain, estimate the market value of the annotation process, give conclusions, and argument on to what extent should future researchers devote their time and money into detailing the annotation granularity of their datasets. This is an important contribution to the community, as we test the feasibility of devoting more time and money on more detailed annotations to increase the task performance. Our outcomes will guide future research of the community, especially given that research related to data annotation are usually very costly~\cite{HovyandLavid2010,johnson-etal-2018-predicting}.

\section{Dataset}

\noindent 
In this paper, the synthetic dataset created and released by Suominen et al.\cite{suominen2015benchmarking} is used. This dataset is created in five steps: 1. Create a patient profile. 2. Collaborate with a registered nurse (RN) to create a synthetic but realistic nursing handover dataset. 3. Create a structured handover form. 4. Create a written, structured document with the form in the third step and written, free-form text documents. 5. Create spoken, free-form text documents.

There are three groups of textual features already annotated in the dataset, namely, syntactic, semantic and statistical features. The syntactic features are about general characteristics of the word itself, including the ``Lemma", ``Named Entity Recognition", ``Part of Speech" tags, and so on. The semantic features are closely related to the semantic meaning of the word, which includes information about the top candidates the word is retrieved with, and the top mappings in the Unified Medical Language System. The statistical features are more about the positioning of the word, for example, the location of the word in a document.

We also introduce the additional annotation by the nurse to later draw comparison before and after this annotation is adopted. During annotation, an Australian registered nurse will manually annotate key points to better describe the patient case. Words are colored to represent different \emph{high-level categories}. Here is a fraction of what the dataset looks like:

\emph{\colorbox{blue!40}{Ken} \colorbox{blue!40}{Harris}, \colorbox{blue!40}{71 yrs old} under \colorbox{blue!40}{Dr Gregor}, complained of \colorbox{purple!50}{chest pain}. \colorbox{green!50}{ECG was done} and \colorbox{green!50}{was reviewed by the team}. He was also given some \colorbox{yellow!100}{anginine}.}

For example, ``Harris" is colored blue representing its label (manually by the nurse) \emph{`PatientIntroduction'}. Importantly, \emph{the task is to see how better the sequence labeling performance becomes (or whether it will become) with and without the expert annotation, to infer a finer level ground truth label.} In other words, we evaluate how much better the model becomes (or whether it will become) after the nurse telling it that ``Harris" belongs to \emph{`PatientIntroduction'} than using purely textual features to infer its (finer level) ground truth: \emph{`PatientIntroduction\_LastName'}.


\section{Method}

\noindent In this section, we introduce the architecture of the trained model, namely \emph{Embeddings + Bidirectional Long Short-Term Memory (BiLSTM) + Conditional Random Field (CRF)}. Lastly, evaluation methods of the model performance and annotation cost are introduced.

\subsection{Embeddings}

\noindent In this work, multiple embeddings of different levels (word embeddings and document embeddings) are stacked up.

\textbf{Word Embeddings.} The \emph{GloVe embedding} is a global log-bilinear regression model \cite{pennington2014glove}. The GloVe model property is necessary to generate linear directions of meaning, and the proposed global log-bilinear regression models are appropriate to do so. 
In this work, GloVe embedding is used as a part of stacked embedding to build up the language model for the sequence labeling task.

\emph{Character Embedding} specifically points to the neural architectures that Lample et al. proposed in 2016 \cite{lample2016neural}. It resolves the problem that good sequence labeling, especially named entity recognition systems heavily rely on hand-crafted features as well as domain-specific features to learn effectively from the available supervised and small training corpora. 

State-of-the-art embeddings including \emph{BERT-embeddings} \cite{devlin2018bert}, \emph{ELMo-embeddings} \cite{peters2018deep} and \emph{Flair-embeddings} \cite{akbik2018contextual} are also adopted.

BERT enables the tasks without abundant training resources to benefit from the architecture, such as this work on Synthetic Nursing Handover Dataset, with a limited amount of training set of only 100 clinical records. 

The word vectors of \emph{ELMo embedding} are learned from the internal states in a deep bi-directional language model (BiLM) trained from a large corpus. The vectors are derived from a BiLSTM trained with a coupled language model based on a large text corpus. This is the reason why ELMo embedding is used in this paper. 

\emph{Flair embedding} \cite{akbik2018contextual}, is a novel embedding pre-trained on a large unlabeled corpora, capturing word meanings in the surrounding context. This makes the model able to produce different embeddings for polysemous words based on their usage.

\textbf{Document Embeddings.} We further use Transformer Document Embeddings, which encodes the whole text piece into a representation. 
 
Transformer Document Embeddings uses the power of transformer, which is already shown in the elaboration of BERT model \cite{devlin2018bert}.We use the robust pre-trained BERT transformer to embed each piece of clinical record into a single representation.

\textbf{Stack Embeddings.} Stack embedding is one of the most important steps used in this project. This means different word embeddings are stacked together to give a more robust embedding~performance. 

For example, we want a combination of the classical embeddings and advanced embeddings. What we do is to use the stack embedding feature to concatenate the two vector representations of the same token, as shown below in Equation (\ref{eq:stack}). 

\begin{equation}
\mathrm{Vec_{stacked}}=[\mathrm{Vec_{glove}} \quad \mathrm{Vec_{bert}}].
\label{eq:stack}
\end{equation}

\subsection{Long Short-Term Memory \& Conditional Random Field}

\noindent This section introduces the overall architecture of the BiLSTM + CRF part in this work. This is to simulate the state-of-the-art performance of current sequence labeling practice. The work mainly focuses on the comparison under the same architecture, rather than the architecture itself.

The Long Short-Term Memory network is a Recurrent Neural Network (RNN) with special architecture. Furthermore, the \emph{Bidirectional} Long Short-Term Memory model is introduced \cite{huang2015bidirectional}. Rather than using a single hidden state $h_t$, which only takes context from the past, the model presents each sequence both forward and backward to two separate hidden states to capture past and future information respectively. The final output of the model is the concatenation of two hidden states.

CRF is a model used for sequence labeling, which has all the advantages of Maximum entropy Markov models, while also solved the problem of label bias \cite{bottou1991approche}. The advantage of CRF is that it can be generalized easily to other kinds of stochastic context-free grammars that are significant in classification tasks in natural language processing, such as the task in this work.

In this work, the BiLSTM + CRF pipeline is used, giving the best performances \cite{chin2018my}.

\subsection{Evaluation Methods}

\noindent This section introduces metrics used in our work to measure the model performance and cost, namely confusion matrix, micro and macro-averaged scores for performance evaluation and return on investment (ROI) for annotation cost estimation.

\paragraph{Confusion Matrix}

The confusion matrix used in this work is a standard version, consisting of the 2x2 grid with True Positive (TP), False Positive (FP), False Negative (FN) and True Negative (TN). We also use the four derived metrics, namely \emph{accuracy}, \emph{precision}, \emph{recall} and \emph{F-1 score} which are given as follows: Accuracy = (TP+TN)/(TP+TN+FP+FN), Precision = (TP)/(TP+FP), Recall=(TP)/(TP+FN), $F_1$=2$\cdot$ (Precision$\cdot$ Recall)/(Precision+Recall).

\paragraph{Macro- and Micro-averaged Evaluation}

To generalize the evaluation above to a wider range of tasks, especially to multi-class classification, we introduce micro and macro-averaged~evaluation. 

For multi-classification tasks, there are generally multiple categories, with each instance belonging to one of them. For a certain class $i$, we denote the metrics as follows: $TP_i$: True positive for class $i$, where the instance belongs to $i$ and is actually classified as $i$. We give the definition in a similar manner for $TN_i$, $FP_i$ and $FN_i$. On top of the four definitions, we can give the definition of $Precision_i$ and $Recall_i$ for each class~$i$ same as above.

We further introduce the micro- and macro-averaged evaluation. Assume there are total $n$ classes in the task, for micro-averaged, we give a representation of Micro-Precision, then Micro-Recall and Micro-$F_1$ can be derived in the same manner:

\begin{equation}\text { Micro-Precision }=\frac{\sum_{i}^{n} \mathrm{TP}_{i}}{\sum_{i}^{n}\left(\mathrm{TP}_{i}+\mathrm{FP}_{i}\right)}.\end{equation}

In addition to the analysis per class, we further have evaluation across all classes using the arithmetic mean, which is known as macro-averaged as~follows (we give macro-averaged precision as an example):

\begin{equation}
    \text{Macro-Precision} =\frac{\sum_{i}^{n} \text { Precision }_{i}}{n}.
\end{equation}

The metrics cover both the overall performance (macro-averaged, the majority classes and those classes with very few occurrences) and performance overall samples (micro-averaged, emphasizing the importance of those dominant majority classes).  

\subsection{Annotation Cost Evaluation}

\paragraph{Return on Investment}

\noindent Return on Investment (ROI) is a financial metric that is widely used to measure the probability of gaining a return from an investment \cite{beattie2020how}.
We use ROI to directly calculate the cost for annotation in this work.

\paragraph{Annotation Cost Rate} \label{acr}

\noindent According to a scaled and normalized estimation standard on ``Scale'', a data platform for AI, the annotation cost for English-language transcription is US\$ 0.08 per token \cite{scale}.

These are the core cost evaluation formula for this work.

\section{Result}

\noindent The result analysis is organized by comparing the model without nursing knowledge and the model with nursing knowledge (i.e. before and after the additional annotation).

\subsection{Annotation without Nursing Knowledge}

\noindent We first look into the model without using the additional manual annotation under the supervision of a registered nurse. The model is studied using pure syntactical, semantic, and statistical knowledge of the clinical trial record.

By first drawing the confusion matrix, we observe that the model is good at identifying classes with a large amount of instances.

\begin{table}[h]
    \centering
    \begin{tabular}{|l|r|r|r|r|r|}
         \hline
    \textbf{Category}  & \textbf{\# Words}& \textbf{TN} & \textbf{FP} & \textbf{FN} & \textbf{TP} \\
      \hline\hline
      \dots & \dots & \dots & \dots & \dots & \dots\\
      \hline
      \textbf{E. NA} & \textbf{3152} & \textbf{3574} & \textbf{1004} & \textbf{395} & \textbf{2757} \\
      \hline
      (1) NA & 3152 & 3680 & 898 & 408 & 2744\\
      \hline\hline
      \textbf{F. PatientIntroduction} & \textbf{2221} & \textbf{82587} & \textbf{225} & \textbf{889} & \textbf{1329}\\
      \hline
      (1) Gender & 544 & 7040 & 146 & 178 & 366\\
      \hline
      (2) Ageinyears & 281 & 7442 & 7 & 1 & 280\\
      \hline
      \dots & \dots & \dots & \dots & \dots & \dots\\
    \hline
    \end{tabular}
    \vspace{2pt}
    \caption{A part example of the handover tag table.}
    \label{tab1}
\end{table}

There are two levels of sequence tagging, for instance, \emph{'PatientIntroduction'} (macro) and \emph{'PatientIntroduction\_Gender'} (micro) (see Table \ref{tab1}).

By analyzing  micro-averaged score, it can be observed that on this multi-class sequence labeling task, the accuracy of the model is extremely high, reaching 0.999 in some classes. It can also be argued that it is right because there are only a few instances in some classes, that the F-1 scores for them are low. 

From the micro-averaged scores, we argue that the model without additional annotation has a mediocre ability to distinguish text contents regarding \emph{`Future'} of the patients, \emph{`Medication'} used, and \emph{`MyShift'} describing the daily life of the patients. But its performance is poor in describing \emph{`Appointment/Procedure'} which refers to a medical appointment with doctors.

\begin{figure}[h!]
    \centering
    \includegraphics[width=\linewidth]{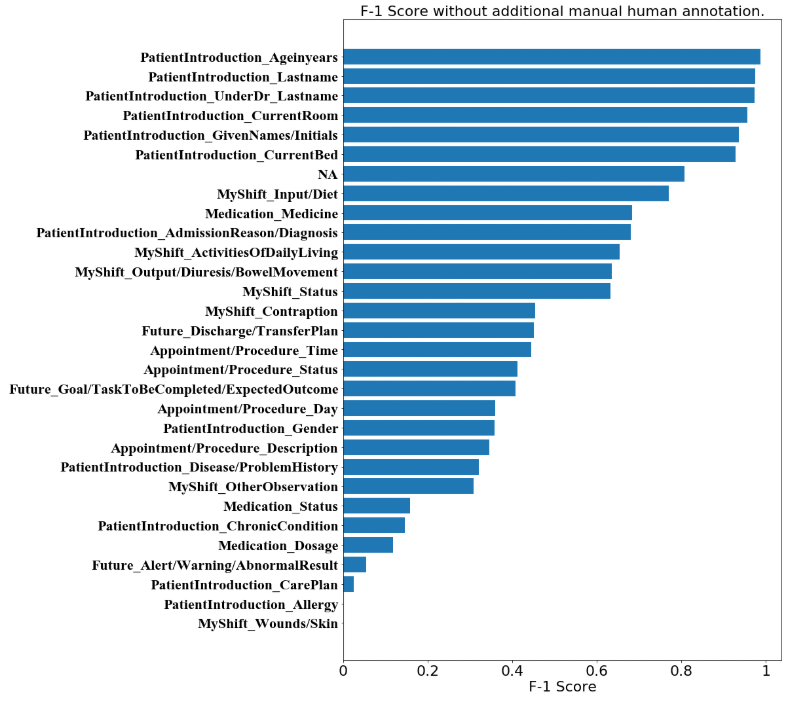}
    \caption{The F-1 score distribution of different categories for the model using only syntactic, semantic, and statistical features.}
    \label{fig:f1.1}
\end{figure}

We then present a straightforward visualization of the F-1 score distribution among all sub-categories in Figure \ref{fig:f1.1}.
From the distribution, we argue that the model without nursing-knowledge-based annotation has an outstanding ability in predicting tags for tokens about the patient introduction, as well as tokens without significant meanings (the \emph{`NA'} tags). 

For the overall macro-averaged score of the model without nursing-knowledge-based annotation, the model performs well in terms of macro-averaged accuracy, which reaches beyond 0.95. On the other hand, it gives a mediocre performance on precision, recall, and F-1 score, with scores near 0.55.

\subsection{Annotation with Nursing Knowledge}

\noindent Similarly, we first draw the confusion matrix of the model trained with nursing knowledge. From the new confusion matrix table, we observe that the four metrics do not seem to differ from those for the model without nursing-knowledge-based annotation. However, it can be observed that the number of predicted labels has changed. In \emph{`Appointment/Procedure'}, \emph{`ClinicianLastname'}, \emph{`ClinicianTitle'} and \emph{`Hospital'} no longer exist, as well as \emph{`RiskManagement'} under \emph{`MyShift'} and \emph{`Allergy'} under \emph{`PatientIntroduction'}. This is not a surprise as these sub-categories were already predicted poorly before adding further manual annotation from the nurse.

 In terms of micro-averaged evaluation scores, the model after adding additional manual annotation is, surprisingly, no much difference to that before adding additional annotations. This is an interesting finding that is against our intuitive idea. This is because adding additional annotation by nursing knowledge means telling the model which high-level category the token belongs to. The model only needs to distinguish the sub-category within the range of the high-level category. However, according to the experiment result, we cannot observe a substantial difference or performance increase except for some minor substantial changes in categories with very few sample sizes, where fluctuation is more likely to occur.

In the sub-categories which remain at the same level, we specifically point out that some of them are consistently distinguished well. This includes \emph{`Input/Diet'} and \emph{`NA'} which reaches a F-1 score near 0.8. This also includes \emph{`Ageinyears'}, \emph{`CurrentBed'}, \emph{`CurrentRoom'}, \emph{`GivenNames/Initials'}, \emph{`Lastname'} and \emph{`UnderDr\_Lastname'}, which reaches F-1 scores near or over 0.95 which are outstanding. It can be argued that 
the model with the stacking of the state-of-the-art embeddings has a good ability to distinguish the tags about names, genders etc. Sub-categories with more than 0.9 F-1 score are mainly under the \emph{`PatientIntroduction'} category.

\begin{figure}[h]
    \centering
    \includegraphics[width=\linewidth]{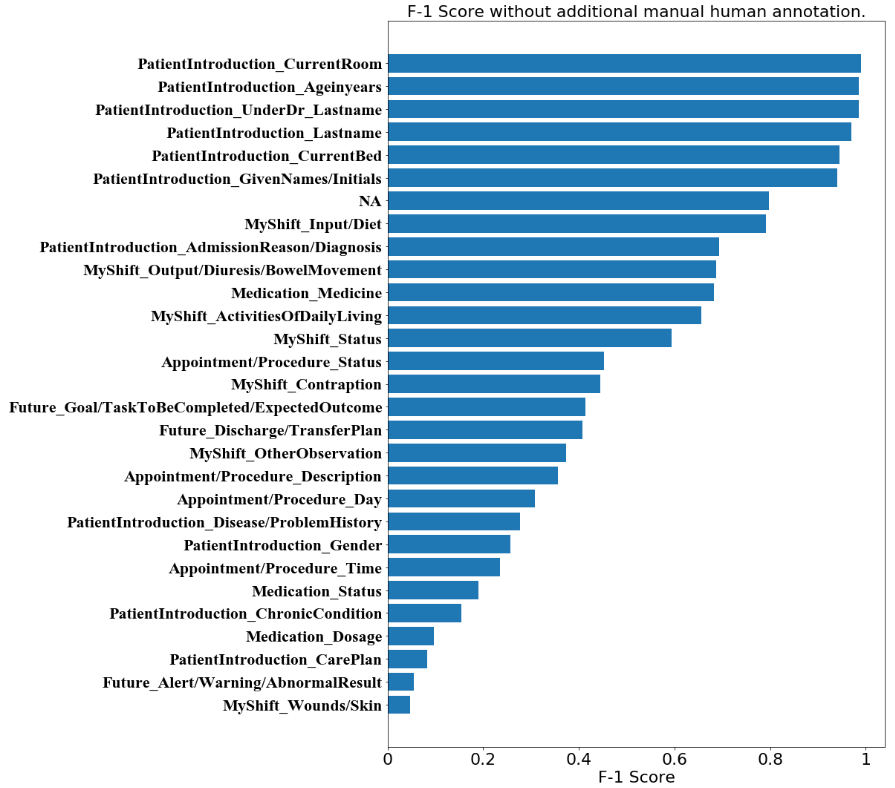}
    \caption{The F-1 score distribution after adding additional manual annotation from the registered nurse.}
    \label{fig:f1.2}
\end{figure}

We visualize the F-1 score distribution for the model after additional manual annotation.
It can be observed that the distribution is no distinct difference to that before the additional annotation. This is both in terms of the absolute values of each sub-category and in terms of the placing among the sub-categories. We observe that most sub-categories under \emph{`PatientIntroduction'} have the highest F-1 scores in both Figures \ref{fig:f1.1} and \ref{fig:f1.2}, with \emph{`NA'} following them in the 7th place. Sub-categories without many instances are those with the lowest F-1 scores. This is reasonable due to insufficient learning by the deep language model.

In terms of macro-averaged score, the model, which includes additional manual annotation does not seem to perform differently from that without the additional manual annotation as well. This is also against the intuitive idea from people that ``adding high-level tag annotation does help the model to classify sub-tags better''. The macro-averaged accuracy, macro-averaged recall, and macro-averaged F-1 score all remain within a 1\% range of fluctuation, while macro-averaged precision increases by 2\%, which is also hard to be argued as a substantial increase given the amount of cost paid into the additional annotation.

\subsection{Visualized Comparison and Cost~Analysis}

\begin{figure}[h]
    \centering
    \includegraphics[width=0.8\linewidth]{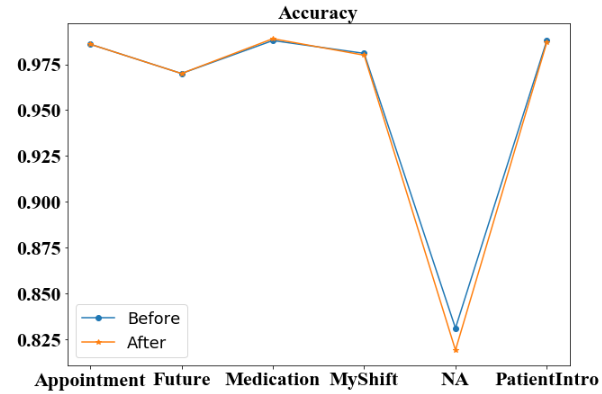}
    \caption{The comparison of the micro-averaged accuracy score for each high-level category classification task before and after the clinical-knowledge-based additional manual annotation.}
    \label{fig:acc_compare}
\end{figure}

\begin{figure}[h]
    \centering
    \includegraphics[width=0.8\linewidth]{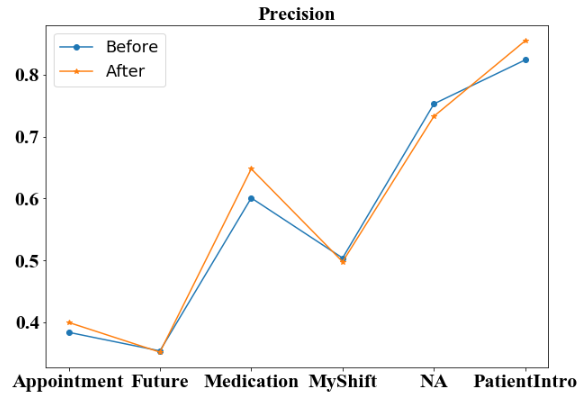}
    \caption{The comparison of the micro-averaged precision score for each high-level category classification task before and after the clinical-knowledge-based additional manual annotation.}
    \label{fig:prec_compare}
\end{figure}

\begin{figure}[h]
    \centering
    \includegraphics[width=0.8\linewidth]{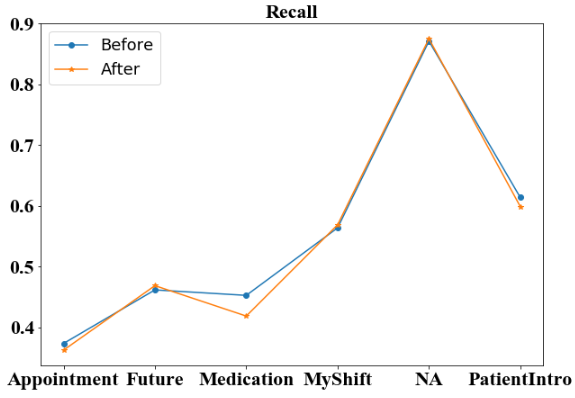}
    \caption{The comparison of the micro-averaged recall score for each high-level category classification task before and after the clinical-knowledge-based additional manual annotation.}
    \label{fig:recall_compare}
\end{figure}

\begin{figure}[h]
    \centering
    \includegraphics[width=0.8\linewidth]{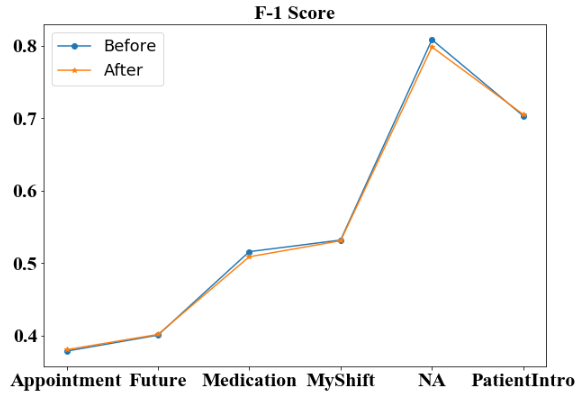}
    \caption{The comparison of the micro-averaged F-1 score for each high-level category classification task before and after the clinical-knowledge-based additional manual annotation.}
    \label{fig:f1_compare}
\end{figure}

\noindent We first compare the four metrics of high-level categories before and after additional annotation.
For three of the four micro-averaged metrics: accuracy (Figure \ref{fig:acc_compare}),  recall (Figure \ref{fig:recall_compare}) and F-1 score (Figure \ref{fig:f1_compare}), the additional annotation does not bring any substantial changes to the performance in most of the 6 high-level categories. However, the additional annotation does yield a worse accuracy score on \emph{`NA'} tag and a large drop in \emph{`Medication'} category from 0.453 to 0.419. But for micro-averaged precision (Figure \ref{fig:prec_compare}), there are noticeable performance increases in 3 of the 6 categories. Specifically, the micro-averaged precision for \emph{`Medication'} class receives a nearly 5\% boost, from 0.601 to 0.648.

\begin{figure}[h]
    \centering
    \includegraphics[width=0.7\linewidth]{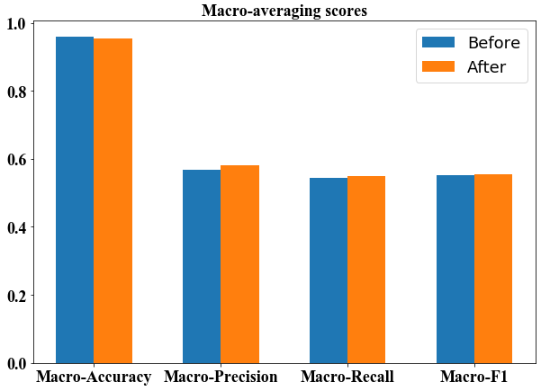}
    \caption{The comparison of macro-averaged scores between the model before and after additional annotation.}
    \label{fig:macro_compare}
\end{figure}

We then compare the macro-averaged scores before and after additional annotation.
From the bar plot (Figure \ref{fig:macro_compare}) we observe that the macro-averaged accuracy of the two models are both high above 0.9, with the 'After' model even being lower. For the macro-averaged precision score, the 'After' model increases for around 1\%, still remaining at a level around 0.55-0.60. For the macro-averaged recall, and macro-averaged F-1 score, we observe that the model before and after additional annotation does not provide any additional information, and both remain at the level around 0.55--0.60.

We finally estimate the human labor annotation cost by a quantitative analysis.
The statistical figures on the Synthetic Nursing Handover Dataset show that the training set and validation set has 8487 tokens annotated in total, while the test set has 7730 tokens annotated. This annotation work is all done by a human Australian registered nurse, where we yield that $N_A = 8487 + 7730 = 16217$. 
According to the sequence labeling result above, we have $F_1^{\prime} = 0.554$ and $F_1 = 0.551$, thus $F_1^{\prime} - F_1 = 0.003$.
According to the parameters calculated above, and the formula for human labor cost rate and return on investment (Equation \ref{eq:roi3}),

\begin{equation}
    R O I=\frac{F_1^{\prime} - F_1}{0.08 \times N_A} \times 100 \%.
    \label{eq:roi3}
\end{equation}

\noindent Thus, we have $ROI = 0.00023124 \%,$
which  
is an extremely small ROI rate. This means that the return for our additional annotation is small enough to be omitted. The estimated pay to the nurse for annotating the dataset is $0.08 \times N_A = $US$\$ \space 1297.36$, where the source of the 0.08 rate is mentioned in Section \ref{acr}. We further calculate that if we want the F-1 score for the model to increase 1\%, we have to pay $1297.36/(1/0.3) = $US$\$ \space 4324.53$. This is not a decent amount of human labor cost in terms of the increase of sequence labeling performance, thus we argue that the additional nursing-knowledge-based annotation in the task is costly, and researchers should not include this kind of additional work when processing their own datasets in medical NLP; instead, we encourage investigating more frugal approaches to use the human time where it is needed the most and developing better NLP workflows to harvest the potential of smaller annotated resources.


\section{Discussion}

\noindent In this work, we researched into the dataset annotation process, which is vital for most of the machine learning and deep learning tasks. We specifically focused on  the annotation granularity and related performance changes and human labor costs.
First, we found that using only syntactic, semantic, and statistical annotation features of the text corpus with well-designed embeddings can yield a good performance in categories with a large amount of instances in the sequence labeling task. Second, we added knowledge-based annotations by an Australian registered nurse into consideration, and found that the text labeling performance remained nearly the same. This is the most surprising finding that is against people's intuitive ideas. This means that adding specialized human knowledge --- even carefully quality-controlled expert annotations --- does not always yield a performance gain in sequence labeling tasks. Third, we formulated the cost of annotation in terms of human labor cost, and derived the conclusion that doing the manual annotation on clinical records is extremely costly in terms of the performance gain of the model. Future researchers in the clinical NLP  should not --- without careful consideration of the anticipated ROI --- spend their time and money investing in dataset annotation in a more detailed~granularity.

However, the work has limitations in the following aspects.
First, the work is very specific in the field of medical NLP, and even specifically for the sequence labeling tasks in the field. Out of the strict attitude to this work, we point out that it cannot be guaranteed that the same conclusion regarding the relationship between classification result and human labor cost can be generalized to other fields of NLP, say, semantics parsing or social media analysis. 
Second, due to the black-box nature of deep learning, we cannot provide a solid explanation of why the detailed expert annotations cannot help the sequence tagger to perform better. However, we can try to control other variables, like try using more and less complex models to study the impact of the underlying embedding complexity and network architecture as future works.


One of the most important contributions of this work is the advice and guideline to the medical NLP researchers in the community. According to our work, we recommend that future researchers and practitioners 
emphasize the linguistic as opposed to medical knowledge as a cost-effective source for increasing the sequence labeling performance with additional expert-annotation in a specific field
when designing, developing, and using expert-annotated
human language technologies and related data resources.
Although capturing medical knowledge is valuable,
expert annotations of this kind are hard and costly to obtain and modify as NLP systems call for revisions and extensions~\cite{HovyandLavid2010,johnson-etal-2018-predicting}.
Hence, we advocate the community to focus more on the pure textual language features. This is because the extra gain from domain-specific knowledge may become marginal as a better textual language feature is extracted from general corpus, especially given the extremely high labor cost for domain-specific knowledge.
\bibliography{acl2020}
\bibliographystyle{IEEEtranS}
\end{document}